\def\year{2017}\relax
\documentclass[letterpaper]{article}
\usepackage{aaai17}
\usepackage{times}
\usepackage{helvet}
\usepackage{courier}

\usepackage{amsthm}
\usepackage[utf8]{inputenc} 
\usepackage[T1]{fontenc}    
\usepackage{hyperref}       
\usepackage{url}            
\usepackage{booktabs}       
\usepackage{amsfonts}       
\usepackage{nicefrac}       
\usepackage{microtype}      

\usepackage{amsmath, amssymb}
\usepackage{times}
\usepackage{helvet}
\usepackage{courier}
\usepackage{amsmath}
\usepackage{multirow}
\usepackage{array}
\usepackage{subcaption}
\usepackage{graphicx}
\usepackage{algorithmicx}
\usepackage[ruled,vlined,linesnumbered]{algorithm2e}
\usepackage{enumitem}
\usepackage{xcolor}

\usepackage[american]{babel} 

\newtheorem{Definition}{Definition}
\newtheorem{Lemma}{Lemma}

\newtheorem{Conjecture}{Conjecture}
\newtheorem{Corollary}{Corollary}
\newtheorem{Conclusion}{Conclusion}
\newcommand{\usc}{\mathcal{C}_{\mathrm{U}}}
\newcommand{\score}{\tau}

\newcommand{\pheadA}[1] {\noindent\textbf{#1.}~} 
\newcommand{\pheadB}[1] {\vspace{1mm}\noindent\textbf{#1.}~} 

\newcommand{\shortening}[1] {}

\begin{document}
%
\title{Should Algorithms for Random SAT and Max-SAT be Different?}

\author{Sixue Liu\\
Microsoft Research \\
Redmond, WA, USA\\
\And
Gerard de Melo\\
Rutgers University\\
New Brunswick, NJ, USA\\
}
\maketitle

\begin{abstract}
    We analyze to what extent the random SAT and Max-SAT problems differ in their properties. Our findings suggest that for random $k$-CNF with ratio in a certain range, Max-SAT can be solved by any SAT algorithm with subexponential slowdown, while for formulae with ratios greater than some constant, algorithms under the random walk framework require substantially different heuristics.
    In light of these results, we propose a novel probabilistic approach for random Max-SAT called ProMS. Experimental results illustrate that ProMS outperforms many state-of-the-art local search solvers on random Max-SAT benchmarks.
\end{abstract}

\section{Introduction}
Given a number of Boolean variables and constraint clauses in conjunctive normal form (CNF), the classic SAT problem consists in determining whether an assignment exists that satisfies all constraints. Its optimization variant, Max-SAT, in contrast, aims to find an assignment that maximizes the number of satisfied clauses.
Unfortunately, both problems are NP-hard, and for the Max-3-SAT variant, it is hard to approximate optimal solutions within a factor of more than $7/8$~\cite{haastad2001some}. This can also be viewed from the perspective of a worst-case upper bound analysis: The time complexity for Max-$k$-SAT (i.e., Max-SAT with clauses of up to $k$ variables) on $n$ variables is $\tilde{O}(2^{(1-\mu)n})$. Although an asymptotic lower bound for $\mu$ has been proven, for $k>2$, no specific constants are known \cite{DBLP:journals/tcs/Williams05,DBLP:conf/sat/ChenS15}.

In the context of random formulae, which have also been the subject of extensive study, the main focus is on the satisfiability threshold and algorithmic polynomial upper bounds for low ratios of clauses to variables \cite{DBLP:journals/siamcomp/AchlioptasM06,DBLP:journals/siamcomp/Coja-OghlanF14}.
Recent theory shows that there exist thresholds for satisfying any large constant fraction of clauses \cite{DBLP:journals/jacm/AchlioptasNP07}.
However, to the best of our knowledge, there is no algorithm tailored to random Max-$k$-SAT.

Algorithms for SAT and Max-SAT are often categorized as either complete or incomplete.
The former guarantee the optimality of the output but may fail to deliver any solution at all within a given time.
Incomplete algorithms, in contrast, can deliver a solution within a given time but do not guarantee its optimality.

Within the paradigm of incomplete algorithms, a common strategy is to rely on local search. Given an objective function, local search algorithms begin with a candidate assignment and traverse the candidate space by iteratively moving to neighboring ones, while keeping track of the best solution encountered in the course of this traversal \cite{DBLP:books/mk/HoosS2004}. If the objective is to minimize the number of unsatisfied clauses, then the incremental nature of such a search implies that local search algorithms for Max-SAT can essentially be used to solve SAT, while SAT algorithms are empirically expected to deliver somewhat reasonable solutions when applied to Max-SAT problem instances.

A number of  local search algorithms have been proposed.
For SAT, the influential GSAT and WalkSAT algorithms pursue a local search in a greedy or probabilistic manner~\cite{Selman1992,selman1994noise}.
More recently, configuration checking (CC)~\cite{cai2012configuration} and probSAT~\cite{balint2012choosing} have made important advances, scoring highest in the random tracks at the SAT competitions 2012 and 2013, respectively.
%
Additionally, a  breakthrough on random $k$-SAT at the phase transition point has been made by polyLS~\cite{sixue2016polyls}.
For Max-SAT specifically,
Iterated Robust Tabu Search~\cite{smyth2003iterated} was ranked first in the Max-SAT Evaluation 2012.
Later, a variant of CC named CCLS~\cite{luo2014ccls}
was highest-ranked in the Unweighted Random track of the Max-SAT Evaluations 2013 and 2015, respectively, improving considerably over previous results.


\pheadB{Contributions}
Our main contribution is to shed new light on the different nature of SAT and Max-SAT when faced with random formulae with different ratios.
%
This theory also has practical implications bearing on the choice of heuristics employed in local search algorithms. We instantiated and empirically evaluated these in a new probabilistic algorithm, called ProMS (Probability Distribution-based Max-SAT Solving), which is shown to outperform the previous state-of-the-art on random instances. The significance of these results also stems from the observation that, despite the success of purely probabilistic strategies for SAT, to the best of our knowledge, no one has previously succeeded in adopting such a strategy for Max-SAT, due to the somewhat different nature of the problems.

\section{Preliminaries}
Our input is a formula $F=c_1 \wedge \dots \wedge c_m$ in conjunctive normal form (CNF), where the $c_i$ are disjunctive clauses that consist of literals (Boolean variables or their negations) on a set of variables $V=\{x_1,x_2,\dots,x_n\}$. A $k$-CNF formula is a CNF such that each clause contains at most $k$ literals.
Given the number of clauses $m$ and number of variables $n$, a \emph{uniform random $k$-CNF} instance consists of $m$ clauses chosen randomly and independently from all $(2n)^k$ possible clauses, with replacement.
We also define $r=\frac{m}{n}$ as the \emph{ratio} of this formula.
Throughout this paper, ``random" shall always refer to ``uniform random", and  $F_k(n,r)$ shall denote a random $k$-CNF with $n$ variables and $m=rn$ clauses. A complete assignment $\alpha$ is a candidate solution such that each variable has a truth value of 0 (\emph{false}) or 1 (\emph{true}).
The Max-SAT problem consists in finding an $\alpha$ that minimizes the number of unsatisfied clauses, denoted as $MaxSAT(F)$. Similarly, $SAT(F)$ is the problem of finding an $\alpha$ satisfying all clauses in $F$.
We use $\mathcal{O}(T(n))$ to represent $poly(n)\cdot T(n)$ where $poly(n)$ refers to polynomials of $n$.

%
%



\section{Analyzing Random SAT and Max-SAT}
We believe that for $F_k(n, r)$, when the ratio $r$ is within a certain (low) range, there is no need to distinguish algorithms for $SAT(F)$ and $MaxSAT(F)$, i.e., an optimal SAT algorithm is also optimal for Max-SAT, while for high ratio $r$, one is well-advised to adopt different heuristics.

\subsection{Low Ratio Random $k$-CNF}
We first consider the low ratio case. We start with the following lemma, and then explain why the precondition within this lemma is very likely true.
\begin{Lemma} \label{opt}
    If the lower bound of the time complexity of all algorithms on random $k$-SAT ($k>2$) with $n$ variables and ratio greater than some constant $r_k^p$ is $\Delta^n$ ($1< \Delta \le 2$), then,
    given $F_k(n, r)$, if there exists an assignment violating $o(m / \log m)$ clauses, if follows that $MaxSAT(F)$ can be solved in $\mathcal{O}(\min(2, \Delta + \epsilon)^n)$ steps for any $\epsilon > 0$.
\end{Lemma}
\begin{proof}
    If $\Delta = 2$, trivially Max-SAT can be solved in $2^n$ steps.
    Otherwise, we solve $MaxSAT(F)$ in $\mathcal{O}((\Delta + \epsilon)^n)$ steps by i) enumerating all the combinations of violated clauses, ii) deleting them from $F$ (we randomly shuffle the order when enumerating deleted clauses to maintain the uniform randomness of $F'$), iii) checking the satisfiability of the remaining formulae $F'$ using the SAT algorithm in $\Delta^n$ steps.
    The number of combinations of violating $o(m/ \log m)$ clauses is bounded by the following:
    \begin{align*}
        \sum_{i=0}^{o(m / \log m)} {\binom{m}{i}}
        &< o(m /\log m) \binom{m}{o(m / \log m)} \\
        &< o(m /\log m) m^{o(m / \log m)} \\
        &= o(m / \log m) 2^{o(m)} \\
        &= 2^{o(m)} = 2^{o(n)}.
    \end{align*}
    Thus the SAT algorithm is invoked at most $2^{o(n)}$ times. Since $SAT(F')$ can be solved in $\Delta^n$ steps, we obtain that $MaxSAT(F)$ can be solved in $\mathcal{O}((\Delta + \epsilon)^n)$ steps for any $\epsilon > 0$.
\end{proof}

The precondition in Lemma~\ref{opt} is that no algorithm can solve random $k$-SAT ($k>2$) with ratio greater than some constant in subexponential time. Four considerations suggest that this conjecture holds true.
\begin{enumerate}
    \item Random walk solves 3-CNF with ratio smaller than 1.63 in polynomial time, while for ratio greater than 1.63 this still remains unknown \cite{DBLP:journals/siamcomp/AlekhnovichB07}.
    \item The best result for solving random $k$-SAT in polynomial time is on formula with $r \le \rho \cdot 2^k / k$ for some constant $\rho$ ($\rho = \frac{1}{25}$ has been proposed in the literature), while for greater $r$ no such algorithm is known \cite{DBLP:journals/siamcomp/Coja-OghlanF14}.
    \item There are exponential lower bounds for DPLL algorithms on satisfiable random 3-CNF formulae \cite{DBLP:conf/sat/AchlioptasM12}.
    \item The exponential time hypothesis (ETH) for the NPC class is widely considered true \cite{DBLP:journals/jcss/ImpagliazzoPZ01}.
\end{enumerate}
Before continuing, we first give the following lemma, which later contributes to our main results.
\begin{Lemma}\label{intro}
    For any integer $k \ge 2$, there exists a $r_k^+ >0$, such that given $F_k(n, r)$ with $r > r_k^+$, there exists an assignment violating $o(m/ \log m)$ clauses with probability $2^{-\Omega(n)}$.
\end{Lemma}
\begin{proof}
    Let $X_s$ be the number of assignments that satisfy at least $s$ clauses in $F$ and let $X_\alpha$ be the number of clauses that a random assignment $\alpha$ satisfies in $F$.
    We have $\mathbb{E}[X_s]=2^n \Pr[X_\alpha \geq s]$. The event that each clause is satisfied by an $\alpha$ is independent, so assuming that $\tilde{X}_{\alpha}$ denotes $\alpha$ satisfying a random clause in $F$, we find that
    \begin{align}
    \mathbb{E}[X_s]
    &= 2^n \sum_{i=s}^{m}\left( \binom{m}{i} \Pr[\tilde{X}_{\alpha}]^i (1-\Pr[\tilde{X}_{\alpha}])^{m-i} \right) \notag \\
    &= 2^n \sum_{i=s}^{m}\left( \binom{m}{i} \left(\frac{2^k - 1}{2^k}\right)^i \left(\frac{1}{2^k}\right)^{m-i} \right).
    \label{eq:sum}
    \end{align}
    Since $i \ge o(m/\log m)$, this sum is dominated by the first term with $i=s$, thus
    \begin{align*}
        \mathbb{E}[X_s]
        &\simeq 2^n \binom{m}{s} \left(\frac{2^k - 1}{2^k}\right)^s \left(\frac{1}{2^k} \right)^{m-s} \\
        &< 2^n \binom{m}{s} \left(\frac{2^k - 1}{2^k}\right)^s .\\
    \end{align*}
    Defining $s = m - o(m / \log m)$, this becomes:
    \begin{align*}
        \mathbb{E}[X_s]
        &< 2^n \binom{m}{o(m / \log m)} \left(\frac{2^k - 1}{2^k}\right)^{m - o(m / \log m)} \\
        &< 2^n \left(\frac{2^k m}{2^k - 1}\right)^{o(m / \log m)} \left(\frac{2^k - 1}{2^k}\right)^m \\
        &< 2^n 2^{o(m)} \left(\frac{2^k - 1}{2^k}\right)^m \\
        &< 2^n \left(\frac{2^k - 1}{2^k} (1 + \delta) \right)^m.
    \end{align*}
    Here $\delta$ can be any positive constant. We choose $\delta = \frac{1}{2^k}$ to bound $\mathbb{E}[X_s]$ from above:
    \begin{equation*}
        \mathbb{E}[X_s]
        < 2^n \left(1 - \frac{1}{4^{k}} \right)^m \\
        = \left( 2 (1 - \frac{1}{4^k} )^r \right)^n.
    \end{equation*}
    By selecting $r = 1 / \log{\frac{4^k}{4^k - 1}} + \eta$ for some positive constant $\eta$, we have $2 (1 - \frac{1}{4^k} )^r < 1$, and $\mathbb{E}[X_s] = 2^{- \Omega(n)}$. Applying Markov's inequality, we obtain
    $$\Pr[X_s > 0] = \Pr[X_s \ge 1] \le \mathbb{E}[X_s] = 2^{-\Omega(n)}.$$
    Recall that $s = m - o(m / \log m)$. The conclusion follows.
\end{proof}

As a counterpart to Lemma~\ref{intro}, we also give the following conjecture regarding low ratio instances, which is analogous to the Sharp Threshold Conjecture for SAT \cite{friedgut1999sharp}.
\begin{Conjecture} \label{lowratio}
    For any integer $k \ge 2$, there exists a $r_k^- >0$, such that given $F_k(n, r)$ with $r < r_k^-$, there exists an assignment violating $o(m/ \log m)$ clauses with probability $1 - 2^{-\Omega(n)}$.
\end{Conjecture}
Conjecture~\ref{lowratio} is compatible with the Sharp Threshold Conjecture for SAT. Thus, $r_k^-$ and $r_k^+$ are also conjectured to converge to the same value when $k$ goes to infinity.
\begin{Conclusion}
    If Conjecture~\ref{lowratio} holds, then for large enough random $k$-CNF ($k > 2$) $F$ with ratio within a certain range, there exists an assignment violating $o(m / \log m)$ clauses with high probability, so by Lemma~\ref{opt} we know that this implies an optimal algorithm for $MaxSAT(F)$.\footnote{Note that we do not consider specific algorithms but only aim to characterize the existence theoretically.}
\end{Conclusion}

\subsection{High Ratio Random $k$-CNF}


We now turn to our main results, providing intuitions about why algorithms for Max-SAT on high ratio formulae should be different from those for SAT (and Max-SAT) on low ratio formulae, with important ramifications for the design of local search heuristics. The following lemma is a simplified version of an upper bound analysis in previous work \cite{DBLP:journals/jacm/AchlioptasNP07}.
\begin{Lemma}\label{constant}
    For any integer $k \ge 2$, given a positive constant $\lambda < \frac{1}{2^k}$, there exists a $r_k^c >0$, such that given $F_k(n, r)$ with $r > r_k^c$, the probability of there existing an assignment violating at most $\lambda m$ clauses is $2^{-\Omega(n)}$.
\end{Lemma}
\begin{proof}
    Set $s = (1 - \lambda) m$ in Eq.~\ref{eq:sum} (in the proof of Lemma~\ref{intro}). Since $\lambda < \frac{1}{2^k}$, this sum is still dominated by the first term, which is
    \begin{equation*}
        \mathbb{E}[X_{(1 - \lambda) m}] \simeq 2^n \binom{m}{\lambda m} \left(\frac{2^k - 1}{2^k}\right)^{(1 - \lambda) m} \left( \frac{1}{2^k} \right)^{\lambda m}.
    \end{equation*}
    Using the fact that $\binom{m}{\lambda m} \sim 2^{h(\lambda)m}$ (omit the polynomial), where $h(\lambda) = -\lambda \log \lambda - (1 - \lambda) \log (1 - \lambda)$ is the \emph{binary entropy function} \cite{cover2012elements}, we obtain:
    \begin{equation*}
        \mathbb{E}[X_{(1 - \lambda) m}] \simeq \left( 2 \left(2^{h(\lambda)} \frac{2^k - 1}{2^k} \left( \frac{1}{2^k-1} \right)^\lambda \right)^r \right)^n.
    \end{equation*}
    If $r > -1 / \left(h(\lambda) + \lambda \log{\frac{1}{2^k - 1}} + \log{\frac{2^k - 1}{2^k}}\right) + \delta$ for any positive constant $\delta$, we have:
    \begin{equation*}
        \Pr[X_{(1-\lambda)m} > 0] \le \mathbb{E}[X_{(1 - \lambda) m}] = 2^{-\Omega(n)}.
    \end{equation*}
    This is the probability of there existing an assignment that violates at most $\lambda m$ clauses, thus concluding the proof.
\end{proof}
Lemma~\ref{constant} immediately implies the following corollary.

\begin{Corollary}\label{cor}
    For any integer $k \ge 2$, given $F_k(n, r)$ with $r > 2^k \ln 2$, at least $f_k^{-1}(r) m $ clauses are violated by any assignment with probability $1 - 2^{-\Omega(n)}$, where $f_k^{-1}$ is the inverse function of $f_k(\lambda) = -1 / (h(\lambda) + \lambda \log{\frac{1}{2^k - 1}} + \log{\frac{2^k - 1}{2^k}})$ and $h$ is the binary entropy function.
\end{Corollary}

In other words, for large enough high ratio random $k$-CNF, a constant fraction of clauses must be violated.
To understand how this constant fraction of violated clauses influences algorithms for Max-SAT, we first give necessary definitions regarding local search.
\begin{Definition}
	Given a CNF $F$ and a complete assignment $\alpha$ for it,
    the \emph{make value} $m(v)$ of a variable $v$ is the number of clauses in $F$ that will transition from unsatisfied to satisfied after flipping $v$ under $\alpha$,
    while the \emph{break value} $b(v)$ of such a $v$ is the number of clauses in $F$ that will transition from satisfied to unsatisfied after flipping $v$ under $\alpha$.
\end{Definition}
\begin{Definition}
    Given a CNF $F$ and an initial complete assignment $\alpha_0$ for it, a local search algorithm $\mathcal{A}$ on $F$ starts with step 0 and $\alpha_0$, and the assignment in step $t$ is denoted as $\alpha(\mathcal{A}, t)$. We further define $\usc(F, t) = \{c \mid \text{clause $c$ violated in $F$ under }\alpha(\mathcal{A}, t)\}$, and if the context is clear, $\usc$ denotes the set of all the currently violated clauses.
\end{Definition}

Let the \emph{Hamming Distance} $d$ be the number of bits on which the current assignment $\alpha$ disagrees with the optimal assignment $\alpha^*$ (violating a minimal number of clauses). An optimal solution is reached if $d=0$.
Note that if there are multiple optimal solutions, this process could terminate earlier, but does not influence our analysis.
We evaluate the influence of the make value on $d$ probabilistically.

For any variable $v$ with make value $m(v)$, flipping $v$ will satisfy $m(v)$ clauses.
For every such clause $c$, $c$ is violated under $\alpha^*$ with probability $\lambda$ (Corollary~\ref{cor}), and thus $d$ increases by 1 after flipping $v$. Another case is more interesting: with probability $1 - \lambda$, $c$ is satisfied under $\alpha^*$, so with probability at least $\frac{1}{k}$, $d$ decreases by 1 because we satisfy the right literal, and with probability at most $1 - \frac{1}{k}$, we set the wrong literal to true and increase $d$ by 1. Simple calculation shows that:
\begin{align}
    \Pr[d \rightarrow d+1] &\le 1 - \left(\frac{1-\lambda}{k}\right)^{m(v)}, \label{make_upper} \\
    \Pr[d \rightarrow d-1] &\ge 1 - \left(1 - \frac{1-\lambda}{k}\right)^{m(v)}. \label{make_lower}
\end{align}
Let us now consider a random walk on a CNF formula, i.e., a Markov Chain with absorbing state $d=0$ \cite{DBLP:conf/focs/Schoning99}.
Decreasing the expectation of steps of a random walk from the initial state to state $d=0$ requires reducing the right side of (\ref{make_upper}) or enlarging the right side of (\ref{make_lower}). Unfortunately this cannot be done at the same time since both of the right sides increase when $m(v)$ increases.
It has empirically been shown that considering only the break value is highly preferable in SAT algorithms \cite{DBLP:conf/aaai/McAllesterSK97}. Further studies have observed that taking the make value into account hurts the performance of local search for SAT \cite{balint2012choosing}. We provide the following theoretical analysis.

\pheadA{Explanation}
Random walk guarantees that $m(v) \ge 1$ by always choosing variables from a violated clause. If we take the make value into account, comparing to $m(v)=1$,
the upper bound of $\Pr[d \rightarrow d+1]$ increases by $\mathcal{U}(\lambda) = \frac{1-\lambda}{k} - (\frac{1-\lambda}{k})^{m(v)}$,
while the lower bound of $\Pr[d \rightarrow d-1]$ increases by $\mathcal{L}(\lambda) = 1 - \frac{1-\lambda}{k} - (1 - \frac{1-\lambda}{k})^{m(v)}$. Define the cost function as
$g(\lambda) = \mathcal{U}(\lambda) / \mathcal{L}(\lambda)$ ($m(v) > 1$), which essentially means increasing the lower bound of $\Pr[d \rightarrow d-1]$ by $x$ will cost an $x \cdot g(\lambda)$ increment in the upper bound of $\Pr[d \rightarrow d+1]$.

Our key observation here is that for $k \ge 2$ and $0 \le \lambda < 1$, $g(\lambda)$ is a strictly decreasing function for $m(v) > 2$, and constant for $m(v) = 2$.
Recall that SAT is the special case of $\lambda = 0$. We have that Max-SAT with positive $\lambda$ has a smaller cost. In other words, for the same improvement on the lower bound of $\Pr[d \rightarrow d-1]$, Max-SAT costs less with respect to increments in the upper bound of $\Pr[d \rightarrow d+1]$. Hence, flipping variables with higher make values does not hurt $\Pr[d \rightarrow d+1]$ as seriously as for SAT.

\begin{Conclusion}\label{c2}
    For sufficiently large random $k$-CNF $F$ ($k \ge 2$) with high ratio, local search algorithms for $MaxSAT(F)$ should more likely consider make values than algorithms for low ratio random $k$-CNF. Moreover, since higher $r$ imply higher $\lambda$ (Corollary~\ref{cor}) and thus smaller $g(\lambda)$, more weight should be given to variables with high make values.
\end{Conclusion}

\section{ProMS Algorithm}
Our analysis leads us to believe that for Max-SAT, the make value, as well, is a crucial piece of information about a variable.
In other words, if $f(v)$ is a scoring function for choosing variables $v$ to flip, an ideal $f(v)$ should be of the form $f(v)=g(m(v),b(v))$, i.e., depend on both the make and break values of a variable $v$. Moreover, Conclusion~\ref{c2} suggests that for higher ratio formulae, variables with high make values should be afforded a higher probability of being flipped, so $f(v)$ may take the form
    $f(v)=g(\zeta(m(v),r ),b(v))$,
where $\zeta(m(v), r)$ is an increasing function on $r$ with $m(v)$ fixed.

Our ProMS algorithm (Algorithm \ref{alg:proms}) first randomly generates a complete assignment, and then repeatedly picks a variable and flips it, for up to a maximal number of steps $M$. In each step, once a clause has been selected, the incident variables are chosen with probability $p$ according to a distribution function $f$. We then update the current assignment. If the number of unsatisfied clauses is now lower than for the previous best assignment $\alpha^{*}$, we update $\alpha^{*}$ to be the current one.
Ultimately, the best found assignment is returned.

\begin{algorithm}[t]
	\small
	\KwIn{CNF-formula $F$, max. steps $M$}
	\KwOut{An assignment $\alpha^{*}$ of $F$}
	generate a random assignment $\alpha$, \ $\alpha^{*}$ $\leftarrow$ $\alpha$\\
	\For {$step$ $\leftarrow$ \upshape{1} \bf{to} $M$}{
		$c$ $\leftarrow$ $pickClause$($\usc(F,\alpha)$)\Comment{random violated clause}\\
		$\score$ $\leftarrow$ $\sum_{v \in c}{f(v)}$ \\
		\If {$\score$ $>$ $\delta$} {
			\ForEach {$v$ $\in$ $c$} {
				choose $v$ and break the loop with probability $\frac{f(v)}{\score}$ \\
			}
		}
		\Else {
			$v$ $\leftarrow$ a variable in $c$ chosen at random\\
		}
		$\alpha$ $\leftarrow$ $\alpha$ with $v$ flipped\\
		\If {$|\usc(F,\alpha)| < |\usc(F,\alpha^{*})|$} {
			$\alpha^{*}$ $\leftarrow$ $\alpha$
		}
	}
	\Return $\alpha^{*}$
	\caption{ ProMS
		\label{alg:proms}
	}
	\vspace*{-0.1cm}
\end{algorithm}


\subsection{Variable Selection Probabilities}
Following our
earlier analysis,
we select variables $v$ for flipping based on both the make values $m(v)$ and the break values $b(v)$, while $m(v)$ is not used in algorithms like WalkSAT and probSAT for the SAT problem.
The polynomial form for $f(v)$ has been proved to be an appropriate choice both in theory and practice \cite{sixue2016polyls},
so we define $f(v)=m(v)^{\zeta}(1+b(v))^{\eta}$, where $\zeta = \zeta(r)$ is an increasing function of ratio $r$.
%
Our algorithm picks variables $v$ with probability
\begin{equation}
\label{par}
p(v)=\begin{cases}
\frac{f(v)}{\score(c)} & \score(c) \geq \delta\\
\frac{1}{|c|} & \text{otherwise},
\end{cases}
\end{equation}
where $\score(c) = \sum_{v \in c} f(v)$ denotes the score of a clause $c$ and $\delta$ is a threshold parameter.

\pheadA{Explanation}
When $\score(c)$ is very low, which implies that all incident variables have high break values or low make values and thus there are no promising variables, 
every variable within the clause is chosen with equal probability. 
Such a purely random selection is also used for diversification in dynamic local search~\cite{li2005diversification}.
Due to the influence of $m(v)$ on $f(v)$, our algorithm could fall into local optima much faster than with a break-only function. Thus the purely random mode
serves to neutralize excessive greediness and prevent our algorithm from performing poorly.
When $\score(c)$ is above the threshold, every variable is allowed to flip with a probability greater than 0 (note that $m(v)$ is always positive because we choose variables from an unsatisfied clause), while in WalkSAT, some flips are forbidden when 0-break variables exist.
Specific choices for the three parameters $\eta$, $\zeta$ and $\delta$ are given in Section~\ref{sec:experiments}.

\section{Experiments}\label{sec:experiments}
We now describe our experiments to assess
ProMS and compare it
with the winners of recent Max-SAT Evaluations.

\pheadB{Parameters}
Our approach has thee parameters: $\eta$, $\zeta$ and $\delta$.
In order to tune these, we use the benchmark data from the Max-SAT Evaluation 2012\footnote{\url{http://www.maxsat.udl.cat/12/benchmarks/index.html}}, because for all of these instances optimal solutions are available\footnote{\url{http://www.maxsat.udl.cat/12/detailed/ms-random-incomplete-table.html}}. 
Based on a grid search,
we set the parameters as $\eta=-2.5$, $\zeta=r+17.5$, and
$\delta=0.4 \cdot r - 1.4$, where $r$ is the ratio.
Note that the $\zeta$ parameter means that for formula with larger ratio, more preference is given to variables according to their make values, which is in line with Conclusion~\ref{c2}.

\subsection{Experimental Setup}

\pheadA{Benchmarks}
We use all of the random 3-CNF\footnote{Random 2-CNF instances in the Max-SAT Evaluation 2016 are too trivial to report, cf. \url{http://maxsat.ia.udl.cat/detailed/incomplete-ms-random-table.html}} instances from the Max-SAT Evaluation 2016 (244 in total), with particularly large ratios ranging from 7.5 to 21.5 to evaluate the robustness of our algorithm. Such high ratio benchmarks are well-suited for an empirical confirmation of our theory, because low ratio formulae do not aid in distinguishing the performance of different solvers in the Max-SAT Evaluation.
The instances are evaluated 20 times each.

\pheadB{Baselines}
We compared ProMS with four state-of-the-art SLS solvers, 
all using the optimal parameters suggested in the referenced literature below.

\begin{itemize}[noitemsep,nolistsep,leftmargin=10pt]
	\item \textbf{probSAT:} We use probSAT, downloaded from EDACC\footnote{\url{http://satcompetition.org/edacc/sc14/experiment/24/solver-configurations/1559}},
	which was the best-performing system in the SAT Competition 2013 ``Sequential Random SAT'' track and the SAT Competition 2014 ``Parallel Random SAT'' track. Parameters are tuned based on the Max-SAT Evaluation 2012, on the same tuning set as for ProMS.
	
	\item \textbf{MaxWalkSAT:} A version of WalkSAT for Max-SAT, obtained from its homepage\footnote{\url{http://www.cs.rochester.edu/u/kautz/walksat/}}.
	
	\item \textbf{iraNovelty++:} The second place in the Max-SAT Evaluation 2013 ``Unweighted Random'' track. We use the latest binary, provided by its author~\cite{abrame2012inference}.
	
	\item \textbf{CCLS:} CCLS~\cite{luo2014ccls} placed first in the Incomplete Solvers track of the Max-SAT Evaluation 2015 ``Unweighted Random'' track. We use the binary submitted to the Max-SAT Evaluation 2015, since the solver for 2016 is not available for download.
	
\end{itemize}

\pheadB{Configurations}
ProMS is implemented in C, and compiled with gcc using the ``-O3'' option for optimization. The cutoff time is set to 300 seconds for all instances and all solvers. All experiments are carried out on a machine with Intel Core Xeon E5-2650 2.60GHz CPU and 32GB RAM under Linux.

\subsection{Evaluation Methodology}

We define the best solution for each instance as the minimal solution found by any of the solvers over all runs. Runs that output the best solution are regarded as successful.
We also define the optimal solution for each solver found for each instance as the minimal one among all runs of the solver on that instance.
We report the average time (``time'') over successful runs, the average optimal solution (``opt.''), and the average solution (``avg.'') over all runs for each class.
A hyphen in the ``time'' columns indicates that a solver failed to deliver a minimal solution in any run.
Please refer to the Max-SAT Evaluation website for further details regarding the
 methodology.\footnote{\url{http://maxsat.ia.udl.cat/detailed/incomplete-ms-random-table.html}}


%



\subsection{Results}



\begin{table*}
	\centering
	\begin{tabular}{|c|c|c|c|c|c|c|c|c|c|c|c|}\hline
		\multirow{2}{1.2cm}[-6pt]{\centering Instance Class} &\multicolumn{2}{c|}{CCLS} & \multicolumn{2}{c|}{iraNovelty++} & \multicolumn{2}{c|}{MaxWalkSAT} & \multicolumn{2}{c|}{probSAT} & \multicolumn{2}{c|}{ProMS} \\
		
		& \multicolumn{1}{m{5.4mm}|}{\centering opt. \\ avg.} &time & \multicolumn{1}{m{5.4mm}|}{\centering opt. \\ avg.} & time & \multicolumn{1}{m{5.4mm}|}{\centering opt. \\ avg.} &  time & \multicolumn{1}{m{5.4mm}|}{\centering opt. \\ avg.} &  time & \multicolumn{1}{m{5.4mm}|}{\centering opt. \\ avg.} &  time \\\hline\hline
		
        v70c700 & \multicolumn{1}{m{5.4mm}|}{\centering 22.8 \\ 23.0} & 2.3 & \multicolumn{1}{m{5.4mm}|}{\centering 23.0 \\ 23.4} & 16.8 & \multicolumn{1}{m{5.4mm}|}{\centering 23.0 \\ 23.0} &  20.8 & \multicolumn{1}{m{5.4mm}|}{\centering 23.2 \\ 23.4} &  4.0 & \multicolumn{1}{m{5.4mm}|}{\centering \textbf{22.8\\ 22.8}} &  \textbf{1.2} \\\hline

    v70c800 & \multicolumn{1}{m{5.4mm}|}{\centering 30.2 \\ 30.4} & 2.1 &
    \multicolumn{1}{m{5.4mm}|}{\centering 30.4 \\ 31.0} & 35.9 &
    \multicolumn{1}{m{5.4mm}|}{\centering 30.6 \\ 31.0} &  38.8 & \multicolumn{1}{m{5.4mm}|}{\centering 31.8\\ 32.4} &  22.1 & \multicolumn{1}{m{5.4mm}|}{\centering \textbf{30.2\\ 30.4}} & \textbf{1.8} \\\hline

    v70c900 & \multicolumn{1}{m{5.4mm}|}{\centering 39.2 \\ 39.4} & 3.9 & \multicolumn{1}{m{5.4mm}|}{\centering 39.4 \\ 40.0} & 10.1 & \multicolumn{1}{m{5.4mm}|}{\centering 40.0 \\ 41.0} &  103.5 & \multicolumn{1}{m{5.4mm}|}{\centering 40.2 \\ 41.4} &  88.3 & \multicolumn{1}{m{5.4mm}|}{\centering \textbf{39.0 \\ 39.4}} &  \textbf{2.0} \\\hline

		v70c1000 & \multicolumn{1}{m{5.4mm}|}{\centering 45.2 \\ 45.4} & 2.9 & \multicolumn{1}{m{5.4mm}|}{\centering 45.6 \\ 45.6} & 21.1 & \multicolumn{1}{m{5.4mm}|}{\centering 45.8 \\ 46.1} &  230.9 & \multicolumn{1}{m{5.4mm}|}{\centering 45.8 \\ 46.6} &  190.3 & \multicolumn{1}{m{5.4mm}|}{\centering \textbf{44.8 \\ 44.8}} & \textbf{2.7} \\\hline
		
		v70c1100 & \multicolumn{1}{m{5.4mm}|}{\centering \textbf{53.8 \\ 53.8}} & \textbf{1.9} & \multicolumn{1}{m{5.4mm}|}{\centering 54.2 \\ 54.4} & 220.8 & \multicolumn{1}{m{5.4mm}|}{\centering 54.0 \\ 54.6} &  105.5 & \multicolumn{1}{m{5.4mm}|}{\centering 54.2 \\ 54.4} &  64.3 & \multicolumn{1}{m{5.4mm}|}{\centering 53.8 \\ 54.0} &  1.3 \\\hline

		v70c1200 & \multicolumn{1}{m{5.4mm}|}{\centering 64.2 \\ 64.4} & 1.8 & \multicolumn{1}{m{5.4mm}|}{\centering 65.0 \\ 65.9} & 76.3 & \multicolumn{1}{m{5.4mm}|}{\centering 65.0 \\ 65.5} &  65.0 & \multicolumn{1}{m{5.4mm}|}{\centering 65.2 \\ 66.0} &  33.4 & \multicolumn{1}{m{5.4mm}|}{\centering \textbf{64.0 \\ 64.2}} &  \textbf{1.5} \\\hline

		v70c1300 & \multicolumn{1}{m{5.4mm}|}{\centering 71.4 \\ 71.6} & 1.5 & \multicolumn{1}{m{5.4mm}|}{\centering 71.6 \\ 72.0} & 10.2 & \multicolumn{1}{m{5.4mm}|}{\centering 72.0 \\ 72.8} &  143.7 & \multicolumn{1}{m{5.4mm}|}{\centering 72.2 \\ 72.4} &  98.3 & \multicolumn{1}{m{5.4mm}|}{\centering \textbf{71.2 \\ 71.4}} &  \textbf{1.5} \\\hline

        v70c1400 & \multicolumn{1}{m{5.4mm}|}{\centering 79.6 \\ 79.6} & 2.9 & \multicolumn{1}{m{5.4mm}|}{\centering 79.8 \\ 80.0} & 33.2 & \multicolumn{1}{m{5.4mm}|}{\centering 79.8 \\ 80.2} &  89.3 & \multicolumn{1}{m{5.4mm}|}{\centering 80.2 \\ 80.6} &  103.2 & \multicolumn{1}{m{5.4mm}|}{\centering \textbf{79.4 \\ 79.6}} &  \textbf{2.3} \\\hline

        v70c1500 & \multicolumn{1}{m{5.4mm}|}{\centering 90.0 \\ 90.2} & 1.8 & \multicolumn{1}{m{5.4mm}|}{\centering 91.2 \\ 91.8} & 45.4 & \multicolumn{1}{m{5.4mm}|}{\centering 90.8 \\ 91.4} &  - & \multicolumn{1}{m{5.4mm}|}{\centering 91.2 \\ 92.0} &  - & \multicolumn{1}{m{5.4mm}|}{\centering \textbf{89.8 \\ 90.0}} &  \textbf{3.7} \\\hline

    v80c600 & \multicolumn{1}{m{5.4mm}|}{\centering 13.4 \\ 13.5} & 2.9 & \multicolumn{1}{m{5.4mm}|}{\centering 13.5\\ 13.5} & 8.5 & \multicolumn{1}{m{5.4mm}|}{\centering 13.5 \\ 13.8} &  99.3 & \multicolumn{1}{m{5.4mm}|}{\centering 13.6 \\ 13.7} &  105.4 & \multicolumn{1}{m{5.4mm}|}{\centering \textbf{13.4 \\ 13.5}} & \textbf{1.9} \\\hline
        		
    v80c700 & \multicolumn{1}{m{5.4mm}|}{\centering 18.8 \\ 18.8} & 3.0 & \multicolumn{1}{m{5.4mm}|}{\centering 19.2 \\ 19.8} & 15.3 & \multicolumn{1}{m{5.4mm}|}{\centering 19.52 \\ 19.8} &  319.3 & \multicolumn{1}{m{5.4mm}|}{\centering 19.4 \\ 19.7} &  404.6 & \multicolumn{1}{m{5.4mm}|}{\centering \textbf{18.7 \\ 18.8}} &  \textbf{2.0} \\\hline

    v80c800 & \multicolumn{1}{m{5.4mm}|}{\centering 27.3 \\ 27.4} & 8.3 & \multicolumn{1}{m{5.4mm}|}{\centering 27.5 \\ 27.8} & 26.3 & \multicolumn{1}{m{5.4mm}|}{\centering 27.4 \\ 27.8} &  155.9 & \multicolumn{1}{m{5.4mm}|}{\centering 27.5 \\ 27.7} &  64.2 & \multicolumn{1}{m{5.4mm}|}{\centering \textbf{27.3 \\ 27.3}} &  \textbf{1.5} \\\hline

    v80c900 & \multicolumn{1}{m{5.4mm}|}{\centering 34.2 \\ 34.2} & 1.9 & \multicolumn{1}{m{5.4mm}|}{\centering 34.4 \\ 34.8} & 45.9 & \multicolumn{1}{m{5.4mm}|}{\centering 34.4 \\ 34.6} &  30.5 & \multicolumn{1}{m{5.4mm}|}{\centering 34.5 \\ 34.6} &  24.4 & \multicolumn{1}{m{5.4mm}|}{\centering \textbf{34.1\\ 34.2}} &  \textbf{2.0} \\\hline

	v80c1000 & \multicolumn{1}{m{5.4mm}|}{\centering 41.0 \\ 41.1} & 3.5 & \multicolumn{1}{m{5.4mm}|}{\centering 41.2 \\ 41.3} & 33.0 & \multicolumn{1}{m{5.4mm}|}{\centering 41.2 \\ 41.3} &  232.9 & \multicolumn{1}{m{5.4mm}|}{\centering 41.2 \\ 41.2} &  14.0 & \multicolumn{1}{m{5.4mm}|}{\centering \textbf{40.9 \\ 41.0}} &  \textbf{1.9} \\\hline

    v90c700 & \multicolumn{1}{m{5.4mm}|}{\centering \textbf{16.9 \\ 17.1}} & \textbf{1.9} & \multicolumn{1}{m{5.4mm}|}{\centering 17.0 \\ 17.5} & 33.6 & \multicolumn{1}{m{5.4mm}|}{\centering 17.1 \\ 17.8} &  260.0 & \multicolumn{1}{m{5.4mm}|}{\centering 17.1 \\ 17.4} &  34.2 & \multicolumn{1}{m{5.4mm}|}{\centering 16.9 \\ 17.1} &  2.1 \\\hline

    v90c800 & \multicolumn{1}{m{5.4mm}|}{\centering 23.3 \\ 23.5} & 4.2 & \multicolumn{1}{m{5.4mm}|}{\centering 23.3 \\ 23.6} & 66.7 & \multicolumn{1}{m{5.4mm}|}{\centering 23.5 \\ 23.8} &  210.0 & \multicolumn{1}{m{5.4mm}|}{\centering 23.5 \\ 24.0} &  - & \multicolumn{1}{m{5.4mm}|}{\centering \textbf{23.1 \\ 23.3}} & \textbf{4.5} \\\hline

    v90c900 & \multicolumn{1}{m{5.4mm}|}{\centering 28.3 \\ 28.5} & 8.3 & \multicolumn{1}{m{5.4mm}|}{\centering 28.5 \\ 28.8} & 98.9 & \multicolumn{1}{m{5.4mm}|}{\centering 28.6 \\ 28.8} &  450.3 & \multicolumn{1}{m{5.4mm}|}{\centering 28.4 \\ 28.6} &  312.2 & \multicolumn{1}{m{5.4mm}|}{\centering \textbf{28.2 \\ 28.3}} &  \textbf{3.5} \\\hline

	v90c1000 & \multicolumn{1}{m{5.4mm}|}{\centering 37.9 \\ 37.9} & 3.5 & \multicolumn{1}{m{5.4mm}|}{\centering 38.4 \\ 38.8} & 119.3 & \multicolumn{1}{m{5.4mm}|}{\centering 38.5 \\ 40.0} &  - & \multicolumn{1}{m{5.4mm}|}{\centering 38.4 \\ 40.1} &  - & \multicolumn{1}{m{5.4mm}|}{\centering \textbf{37.8 \\ 37.8}} & \textbf{2.9} \\\hline

		v90c1100 & \multicolumn{1}{m{5.4mm}|}{\centering 45.3 \\ 45.4} & 5.8 & \multicolumn{1}{m{5.4mm}|}{\centering 45.6 \\ 45.9} & 46.2 & \multicolumn{1}{m{5.4mm}|}{\centering 45.8 \\ 50.1} & - & \multicolumn{1}{m{5.4mm}|}{\centering 46.0 \\ 49.5} & - & \multicolumn{1}{m{5.4mm}|}{\centering \textbf{45.1 \\ 45.2}} &  \textbf{14.2} \\\hline

		v90c1200 & \multicolumn{1}{m{5.4mm}|}{\centering 53.6 \\ 53.7} & 6.4 & \multicolumn{1}{m{5.4mm}|}{\centering 53.9 \\ 54.4} & 33.1 & \multicolumn{1}{m{5.4mm}|}{\centering 54.0 \\ 54.3} & - & \multicolumn{1}{m{5.4mm}|}{\centering 54.1 \\ 54.4} & - & \multicolumn{1}{m{5.4mm}|}{\centering \textbf{53.5 \\ 53.7}} & \textbf{8.3} \\\hline

		v90c1300 & \multicolumn{1}{m{5.4mm}|}{\centering 61.6 \\ 62.1} & 3.1 & \multicolumn{1}{m{5.4mm}|}{\centering 61.9 \\ 62.8} & 102.7 & \multicolumn{1}{m{5.4mm}|}{\centering 62.0 \\ 62.8} & - & \multicolumn{1}{m{5.4mm}|}{\centering 61.8 \\ 62.7} & 332.4 & \multicolumn{1}{m{5.4mm}|}{\centering \textbf{61.4 \\ 61.6}} & \textbf{22.1} \\\hline

    v110c700 & \multicolumn{1}{m{5.4mm}|}{\centering 10.4 \\ 10.6} & 8.2 & \multicolumn{1}{m{5.4mm}|}{\centering 10.5 \\ 10.7} & 40.8 & \multicolumn{1}{m{5.4mm}|}{\centering 10.5 \\ 10.8} &  117.0 & \multicolumn{1}{m{5.4mm}|}{\centering 10.4 \\ 10.8} &  51.6 & \multicolumn{1}{m{5.4mm}|}{\centering \textbf{10.3 \\ 10.5}} & \textbf{11.4} \\\hline

    v110c800 & \multicolumn{1}{m{5.4mm}|}{\centering 16.5 \\ 16.6} & 3.1 & \multicolumn{1}{m{5.4mm}|}{\centering 16.6 \\ 16.7} & 56.4 & \multicolumn{1}{m{5.4mm}|}{\centering 16.7 \\ 16.8} &  100.4 & \multicolumn{1}{m{5.4mm}|}{\centering 16.7 \\ 16.9} &  44.8 & \multicolumn{1}{m{5.4mm}|}{\centering \textbf{16.5} \\ \textbf{16.6}} & \textbf{2.4} \\\hline

    v110c900 & \multicolumn{1}{m{5.4mm}|}{\centering 21.6 \\ 21.7} & 4.1 & \multicolumn{1}{m{5.4mm}|}{\centering 21.8 \\ 22.0} & 30.3 & \multicolumn{1}{m{5.4mm}|}{\centering 21.9 \\ 22.3} &  133.2 & \multicolumn{1}{m{5.4mm}|}{\centering 21.8 \\ 22.0} &  71.0 & \multicolumn{1}{m{5.4mm}|}{\centering \textbf{21.5 \\ 21.7}} & \textbf{9.4} \\\hline

	v110c1000 & \multicolumn{1}{m{5.4mm}|}{\centering 44.9 \\ 45.1} & 1.2 & \multicolumn{1}{m{5.4mm}|}{\centering 45.0 \\ 45.3} & 38.5 & \multicolumn{1}{m{5.4mm}|}{\centering 45.5 \\ 45.8} & 130.4 & \multicolumn{1}{m{5.4mm}|}{\centering 45.6 \\ 45.9} & - & \multicolumn{1}{m{5.4mm}|}{\centering \textbf{44.8 \\ 45.0}} & \textbf{1.2} \\\hline

	v110c1100 & \multicolumn{1}{m{5.4mm}|}{\centering 37.0 \\ 37.4} & 1.9 & \multicolumn{1}{m{5.4mm}|}{\centering 37.8 \\ 38.0} & 313.9 & \multicolumn{1}{m{5.4mm}|}{\centering 38.0 \\ 38.3} & - & \multicolumn{1}{m{5.4mm}|}{\centering 38.1 \\ 38.5} & - & \multicolumn{1}{m{5.4mm}|}{\centering \textbf{36.9 \\ 37.0}} & \textbf{22.5} \\\hline



	\end{tabular}
	\caption{Each line represents a class of instances, with the number of variables and clauses designated in its name, containing many instances of the same size.
The best performance on each class is in \textbf{bold}, defined as the solver with minimal ``opt.'', breaking ties with smaller ``avg.'' and then ``time''.
MaxWalkSAT and probSAT are unable to find the best solution for all instances, while iraNovelty++ finds solutions with better quality but also substantially longer average running time than others.
ProMS gives the solutions with best quality on all classes except \emph{v70c1100}, and on \emph{v90c700} it outputs the same solution with 0.2s slowdown in running time comparing to CCLS.
}
	\label{tab:maxsat2016}
\end{table*}

The experimental results are given in Table~\ref{tab:maxsat2016}. CCLS is based on Configuration Checking and a dynamic clause weighting scheme. Although the latter can be regarded as a form of exploiting the make value, CCLS fails to compete with ProMS, particularly on high ratio instances.
MaxWalkSAT, in contrast, is based on WalkSAT and only considers the break value. Even with parameters tuned specifically for Max-SAT, it lags far behind CCLS and ProMS. This indicates that the make value plays a key role for Max-SAT.

Interestingly, probSAT, which pursues a similar strategy to ProMS, but neglects make values, turns out to be among the weakest of all approaches. Recall that probSAT and its variants represent the state-of-the-art for random SAT. This confirms our conjecture that algorithms for Max-SAT and SAT need to differ in heuristics that they consider.

\section{Conclusions}

\begin{figure}[t]
\centering
\includegraphics[scale=1]{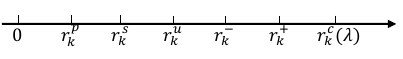}
\caption{For random $k$-CNF, with high probability, ratios below $r_k^p$ have a polynomial decidable algorithm; ratios below $r_k^s$ are satisfiable; ratios below $r_k^-$ have $o(m / \log m)$ violated clauses; ratios beyond $r_k^u$ are unsatisfiable; ratios beyond $r_k^+$ have $\Omega(m / \log m)$ violated clauses; ratios beyond $r_k^c(\lambda)$ have $\lambda m$ violated clauses.}
\label{magnitude}
\end{figure}

Should algorithms for random SAT and Max-SAT be different?
We have attempted to approach this question both theoretically and empirically. The relevant theory (including conjectures) can be summarized by Figure~\ref{magnitude}, where $r_k^-$ and $r_k^+$ are the thresholds proposed in this work. This suggests that for formulae with ratio in the range $[r_k^p, r_k^-]$, there is no need to distinguish between SAT and Max-SAT algorithms, while for ratios higher than $r_k^+$, the nature of the two problems is different. More specifically, our work suggests that under the random walk framework, not only break values but also make values are to be considered when choosing variables to flip, and more preference should be given to the latter when the ratio goes up.

To back up our findings and as an additional contribution,
we have presented a novel algorithm for Max-SAT called ProMS.
Unlike most previous Max-SAT approaches, ProMS eschews a greedy strategy in favor of a more probabilistic one.
Contrasting between algorithms with and without make value, as well between ProMS and probSAT, our empirical findings confirm the value of exploiting
make values and corroborate our theoretical conclusion that algorithms for SAT and Max-SAT should differ for high ratio formulae.

Regarding future work, note that the gap between $r_k^s$ and $r_k^u$ never closes for constant $k$, so exploring a rigorous lower bound for $r_k^-$ and its gap to $r_k^+$ for small $k$ will be an intriguing direction. Further, new theory based on a deeper understanding of the geometry of the solution space may also aid in analyzing and improving the design of heuristics. From a practical perspective, having found that
ProMS outperforms the winners in Max-SAT Evaluations
leads us to consider self-adaptive parameters and clause weighting schemes to apply our ideas to further kinds of Max-SAT instances.
Another interesting point is the strong randomness in the clause and variable selection, which brings considerable variance. Thus engineering a Max-SAT solver incorporating the ProMS algorithm with multiple threads would achieve much better results in practice.

\section{Acknowledgments}
This research was done in part when both authors were at IIIS, Tsinghua University.

\bibliographystyle{alpha}
\bibliography{proms_aaai}

\end{document}